\documentclass[10pt,twocolumn,letterpaper]{article}
\usepackage{iccv}

\usepackage{times}
\usepackage{epsfig}
\usepackage{graphicx}
\usepackage{amsmath}
\usepackage{amssymb}
\usepackage{caption}
\usepackage{bbding}
\usepackage{color}
\usepackage[accsupp]{axessibility} 
\usepackage[hang,flushmargin]{footmisc} 

\usepackage[breaklinks=true,bookmarks=false]{hyperref}

\iccvfinalcopy 

\def\iccvPaperID{8297} 

\ificcvfinal\fi

\begin{document}
	
	\title{Bridging the Gap between Label- and Reference-based Synthesis in Multi-attribute Image-to-Image Translation}
	
	\author{$\text{Qiusheng Huang}^1$, $\text{Zhilin Zheng}^2$, $\text{Xueqi Hu}^1$, $\text{Li Sun}^1$\footnotemark[1], $\text{Qingli Li}^1$\\
		$^1$Shanghai Key Laboratory of Multidimensional Information Processing, East China Normal University\\
		$^2$PingAn Technology}
	
	\twocolumn[{
		\renewcommand\twocolumn[1][]{#1}
		\maketitle
		\begin{center}
			\centering 
			\includegraphics[width=0.80\textwidth]{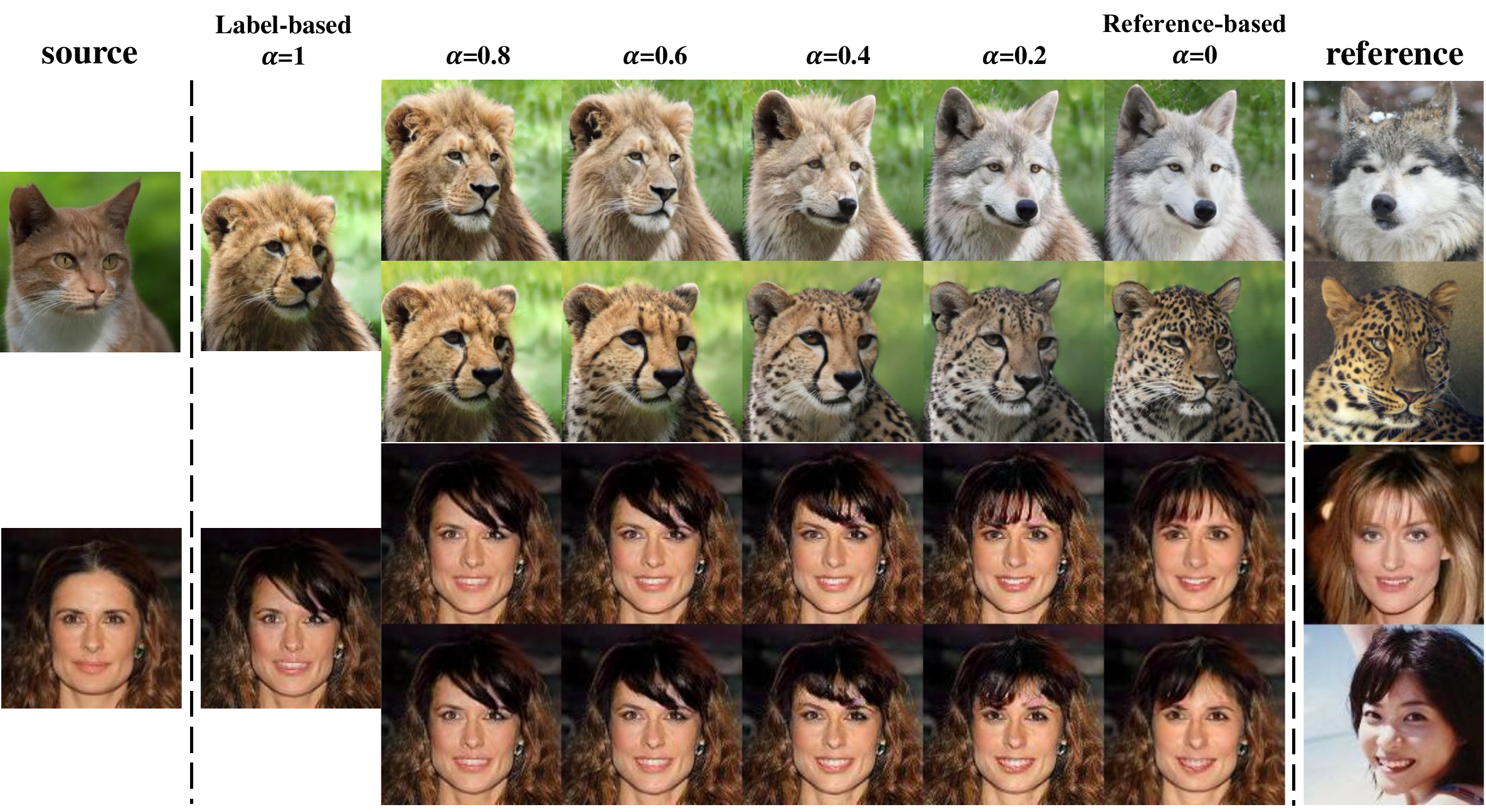}
			\captionof{figure}{\textbf{Interpolation results between label- and reference-based synthesis.} For the species translation, 
				a cat face can be turned into other wild animals. For the face editing, the model converts "bangs" and "mouth open" at the same time. 
				$\alpha$ is the interpolation rate. $\alpha = 1$ means the result is entirely label-based, 
				while $\alpha=0$ takes the style from the reference.
				\label{fig:fig1}
			}
			
		\end{center}
	}]
	\ificcvfinal\thispagestyle{empty}\fi
	
	\begin{abstract}
		The image-to-image translation (I2IT) model takes a target label or a reference image as the input, and changes a source into the specified target domain. The two types of synthesis, either label- or reference-based, have substantial differences. Particularly, the label-based synthesis reflects the common characteristics of the target domain, and the reference-based shows the specific style similar to the reference. This paper intends to bridge the gap between them in the task of multi-attribute I2IT. We design the label- and reference-based encoding modules (LEM and REM) to compare the domain differences. They first transfer the source image and target label (or reference) into a common embedding space, by providing the opposite directions through the attribute difference vector. Then the two embeddings are simply fused together to form the latent code $S_{rand}$ (or $S_{ref}$), reflecting the domain style differences, which is injected into each layer of the generator by SPADE. To link LEM and REM, so that two types of results benefit each other, we encourage the two latent codes to be close, and set up the cycle consistency between the forward and backward translations on them. Moreover, the interpolation between the $S_{rand}$ and $S_{ref}$ is also used to synthesize an extra image. Experiments show that label- and reference-based synthesis are indeed mutually promoted, so that we can have the diverse results from LEM, and high quality results with the similar style of the reference. Code will be available at \textcolor{red}{https://github.com/huangqiusheng/BridgeGAN}. \noindent\footnotetext{Corresponding author, email: sunli@ee.ecnu.edu.cn. This work is supported by the Science and Technology Commission of Shanghai Municipality (No.19511120800). }
	\end{abstract}
	\vspace{-0.2cm}
	\section{Introduction}
	
	Image-to-image translation (I2IT) aims to learn mapping functions among different domains. These domains are either defined by a single attribute \cite{isola2017image,zhu2017unpaired}, therefore, they are mutually exclusive, \emph{e.g.} changing a cat face into a dog. Or they are specified by multiple attributes, so one domain may be overlapped with others with respect to a different attribute, \emph{e.g.} the hair color and the gender are different attributes. Naturally, a domain of black hair has intersection with the male domain. An ideal I2IT model should be able to change the source images into the required target domain, while keeping the content of the source without excessive modifications. For multi-attribute I2IT, since the model needs to complete the translations according to multiple source-to-target requirements, it becomes important it has the ability to accurately edit the individual attribute-related domain, while making other unrelated domains stable. In practice \cite{choi2018stargan,liu2019stgan}, the intended domain labels usually participate the generation, so we name such results the label-based synthesis. 
	
	On the other hand, images are often with various styles even if they are in the same domain. \emph{E.g.}, bangs or beards may look quite different. Therefore, it is expected
	to synthesize multiple results within the same domain, and the styles of them must be under users' control. By providing a reference and asking the model to synthesize an image in the similar 
	appearance with it, we can get diverse multi-modal results \cite{zhu2017toward,huang2018multimodal,lee2020drit++,choi2020stargan}. However, the reference-based synthesis is difficult to be 
	realized in the multi-attribute I2IT. Most of the existing works only deal with a single attribute, which means that one domain is not overlapped with others. Only a few works \cite{chen2019homomorphic,xiao2018elegant,liu2020gmm} aim at the multi-attribute setting, but their results are often poor, compared with the label-based synthesis.
	
\begin{figure*}[ht]
		\centering
		\includegraphics[
		width=0.85\textwidth]{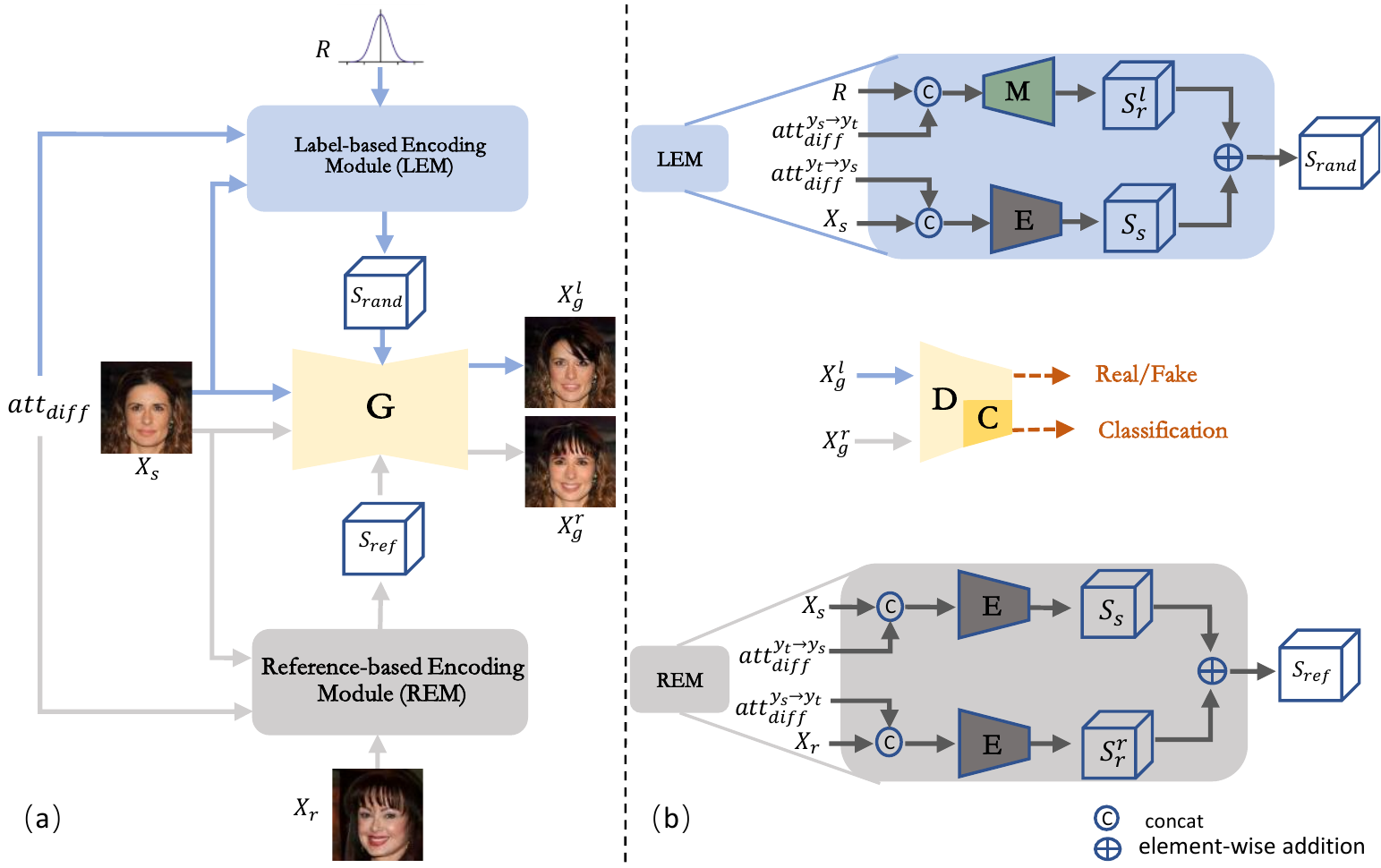}
		\caption{\textbf{The overall structure based on LEM and REM.} On the left, we give the overview of our model. On the right, details about LEM and REM are provided. All encoders (E) share the model parameters.}
		\label{fig:fig2}
	\end{figure*}

	This paper aims to build a single model to bridge the gap between the label- and reference-based synthesis for the multi-attribute I2IT, as is shown in Fig.\ref{fig:fig2}(a). Primarily, our model translates the source image $X_s$ into the target domain, through either the label difference vector $\mathrm{att}_{diff}$, or the reference image $X_r$ lying in a  domain different from the source. The results $X_g^l$ from the former (label-based synthesis) are usually of high quality and in the correct required domain. But they have only a single mode and lack diversity. The latter $X_g^r$ (reference-based synthesis) can potentially generate multi-modal images, but are often of low quality and in the wrong domain. Our idea is to utilize the two types of synthesis, and make them guide each other, so that the final results from both of them get promoted. 
	
	Specifically, we design two units which are Label- and Reference-based Encoding Module, referred as LEM and REM in Fig.\ref{fig:fig2}(a). Their outputs are given to the common main branch of the generator $\text G$ to synthesize $X_g^l$ and $X_g^r$. Both modules have two branches. In REM, they process the source $X_s$ and reference $X_r$, respectively. One branch encodes $X_s$ into a latent code along the direction specified by the difference between source and target domain labels, while the other encodes $X_r$ in the similar way, but in the opposite direction. The two branches intend to find a pair of latent codes from $X_s$ and $X_r$, respectively. And they are all used by $\text G$, which are finally translated into a target domain $X_g^r$, with its style being similar to $X_r$. The LEM mimics the design of REM. It also encodes $X_s$ in the direction of the target domain, in the same way as REM. However, since it has no specific reference as the input, we sample a random noise vector to replace it, and design a separate module to map the noise into the latent code. The results from both branches are fused together, and given to $\text G$ for the $X_g^l$.
	
	To further bridge the performance gap between the results of $X_g^l$ and $X_g^r$, the model not only outputs the two of them, but also a translated image based on the interpolation of the two latent codes from the LEM and REM. Moreover, we design a constraining loss directly on the two codes. Intuitively, both of them are used by the common module $\text G$ to translate the same $X_s$, so they should be close to each other. To better connect the LEM and REM, and encourage the $X_g^l$ to have diverse styles, we explicitly assign a unique noise vector to each reference $X_r$ during training, and minimize the distance between two latent codes from LEM and REM for the same pair of $X_s$ and $X_r$. In addition, $X_g^l$ and $X_g^r$ are fed back into the REM as a reference or a source, therefore, the cycle consistency can be applied. We do extensive ablations on each objective loss terms. Fig.\ref{fig:fig1} shows our impressive visual results. 
	
	
	\section{Related Works}
	\noindent\textbf{I2IT for a single attribute.} The topic of I2IT is first proposed in \cite{isola2017image}. The pix2pix and its later version pix2pixHD \cite{wang2018high} are made of autoencoders, which translate images between two domains based on paired data. CycleGAN \cite{zhu2017unpaired} extends it to the unpaired data. However, these models only generate single-modal results. 
	
	To encourage diverse styles, one way is to change the latent code from deterministic to probabilistic, usually achieved by VAE \cite{zhang2020disentangling, kingma2013auto,bao2017cvae, higgins2016beta, sohn2015learning, esser2018variational, zheng2019disentangling, yin2020novel, chen2018isolating}. BicycleGAN \cite{zhu2017toward} gives the multi-modal results by employing the VAE structure. UNIT \cite{liu2017unsupervised} uses two VAE encoders, mapping images from different domains into a shared space. MUNIT \cite{huang2018multimodal} and DRIT \cite{lee2018diverse,lee2020drit++} disentangle between the content and the style code to encourage the diverse styles. They also add an extra encoder, specifying a style code which is injected into the generator by AdaIN \cite{huang2017arbitrary}. With a similar structure, FUNIT \cite{liu2019few} extends previous works in the few shot scenarios, and the model can work for multiple domains, with each one having only a few examples. StarGAN-V2 \cite{choi2020stargan} and TUNIT \cite{baek2020rethinking} also aim at multi-domain translation, but they can not perform well in the multi-attribute setting.
	
	\noindent\textbf{I2IT for multi-attributes.} Domains defined by different attributes are inevitably overlapped with each other. StarGAN \cite{choi2018stargan} and AttGAN \cite{he2019attgan} are two similar models for label-based synthesis. The source image and the required labels are processed together to give the results. STGAN \cite{liu2019stgan} refines the structure by setting up connections from encoder to decoder. RelGAN \cite{wu2019relgan} uses the label difference to replace the source and target labels. However, results from them are single-modal. To increase the diversity, SMIT \cite{romero2019smit} simply incorporates the random noises on the style code to augment the label-based synthesis, making a tradeoff between the quality and diversity. ELEGANT \cite{xiao2018elegant} and HomoGAN \cite{chen2019homomorphic} are pure reference-based models. Some works like GMM-UNIT \cite{liu2020gmm} and DMIT \cite{yu2019multi} build a single model to support both the label- or reference-based synthesis by introducing the probabilistic encoder. The two types of synthesis are obtained by sampling from the prior or the posterior. Nonetheless, modeling distribution for every domain is difficult particularly when considering large number of attributes. 
	\section{Proposed Method}
	
	\subsection{Problem Formulation}
	Our model aims to translate an image $X_s \in \mathbb{R}^{H\times W\times 3}$, with its multi-attribute binary label $Y_s\in \{0,1\}^n$, into an image $X_g$ in a different domain specified by a target label $Y_t\in \{0,1\}^n$. The reference image $X_r$ is optionally provided during the inference, specifying a particular target domain style for $X_g$. Note that this is a typical unpaired generation task in which we do not have the groundtruth for $X_g$ during training. Here $n$ is the number of the attributes, and each one defines two non-overlapped visual domains, meaning with or without a specific attribute. In total, there are $2^n$ different domains. $\mathrm{att}_{diff}^{Y_s\rightarrow Y_t}\in \{-1,0,+1\}^n=Y_t-Y_s$ is also an $n$ element vector, representing the direction from source to target. It is employed by the LEM and REM as the input condition. Fig.\ref{fig:fig2} illustrates the specific architecture of our model, consisting of a mapping network $\text M$, an encoder network $\text E$, a generator network $\text G$ and a discriminator $\text D$ with an extra multi-attribute domain classifier $\text C$ \cite{odena2017conditional}. The two types of synthesis $X_g^l$ and $X_g^r$ are built on LEM and REM modules, respectively. In summary, given following inputs: an image pair $X_s$ and $X_r$, a noise vector $R$, and two opposite directions $\mathrm{att}_{diff}^{Y_s\rightarrow Y_t}$ and $\mathrm{att}_{diff}^{Y_t\rightarrow Y_s}$, the LEM and REM are designed to output the latent codes for the label- and reference-based synthesis, $X_g^l$ and $X_g^r$.
	
	\subsection{Pipelines for Two Types of Synthesis}
	The two modules, LEM and REM, support the two types of synthesis $X_g^l$ and $X_g^r$ by injecting their outputs $S_{rand}$ and $S_{ref}$ into $\text G$. 
	They essentially compare the two inputs from different domains, and encode their differences into a style code. Note that both modules are composed of two branches, where each branch maps its input into an intermediate latent code, and then they are combined together. These processes are summarized in (\ref{eq:eq1}) and (\ref{eq:eq2}). Details are illustrated in following subsections.
	\begin{equation}
		\label{eq:eq1}
		\begin{aligned}
			S_r^l = \mathrm{M} (R, \mathrm{att}_{diff}^{Y_t\rightarrow Y_s}) \quad S_r^r = \mathrm{E} (X_r, \mathrm{att}_{diff}^{Y_t\rightarrow Y_s}) \\ S_s = \mathrm{E} (X_s, \mathrm{att}_{diff}^{Y_s\rightarrow Y_t})
		\end{aligned}
	\end{equation}
	\begin{equation}
		\label{eq:eq2}
		\begin{aligned}
			S_{rand} = \mathrm{LEM} (X_s, R, \mathrm{att}_{diff})=S_s+S_r^l\\ S_{ref} = \mathrm{REM} (X_s, X_r, \mathrm{att}_{diff})=S_s+S_r^r
		\end{aligned}
	\end{equation}
	
	\indent\textbf{LEM for label-based synthesis.} 
	The mapping network $\text M$ encodes the random noise $R\in\mathbb{R}^d$ together with $\mathrm{att}_{diff}^{Y_t\rightarrow Y_s}$, and gradually increases the spatial size until $S_r^l\in\mathbb{R}^{\frac{H}{k}\times \frac{W}{k}\times C}$. In practice, we concatenate $R$ with $\mathrm{att}_{diff}^{Y_t\rightarrow Y_s}$ before giving it to $\text M$, as is shown in (\ref{eq:eq1}). Similarly, the source $X_s$ is encoded by $\text E$ in the opposite direction of $\mathrm{att}_{diff}^{Y_s\rightarrow Y_t}$ to form another intermediate code $S_s$ with the same size as $S_r^l$. Then, $S_r^l$ and $S_s$ are added together to form $S_{rand}$ like (\ref{eq:eq2}), which reflects the domain style differences on the specified attributes. In terms of purpose, this is equivalent to using the same $\mathrm{att}_{diff}$ to generate $S_s$ and $S_r^l$, and then get $S_{rand}$ by $|S_r^l-S_s|$. The synthesis $X_g^l$ based on $S_{rand}$ shows the common characteristics in the target domain, but it often lacks diversity even if we can sample a random $R$ as the input. So we need $\text {REM}$ to take the effect and give the 
	multi-modal results. Furthermore, we emphasize that incorporating with $\mathrm{att}_{diff}^{Y_s\rightarrow Y_t}$ can help to locate the attributes to be transferred, at the same time, maintain the remaining attributes specified by $0$ in $\mathrm{att}_{diff}$. 
	
	\indent\textbf{REM for reference-based synthesis.} $\text{REM}$ 
	has the same structure as LEM to process $X_s$, mapping it into $S_s$ by $\text{E}$ in the direction of $\mathrm{att}_{diff}^{Y_s\rightarrow Y_t}$. As is shown in (\ref{eq:eq1}), $\text{E}$ also encodes $X_r$ to get a code $S_r^r$. Then, $S_r^r$ and $S_s$ are added to form $S_{ref}$ like in (\ref{eq:eq2}). Note that the result $X_g^r$ based on $S_{ref}$ not only lies in the target domain but also has the similar style as $X_r$. 
	
	\textbf{The generator G for two types of synthesis.} $S_{rand}$ and $S_{ref}$ are employed by a common $\text{G}$ to give the final results $X_g^l$ and $X_g^r$, respectively. $\text{G}$ is an auto-encoder, which first encodes $X_s$ into an embedding space, then decodes it back into an image. $S_{rand}$ and $S_{ref}$ are employed by the main branch of decoder $\text G$ through SPADE \cite{Park_2019_CVPR}. 
	Note that the parameters of $\text G$ and SPADE are shared for $S_{rand}$ and $S_{ref}$.
	
	\subsection{Training Objectives}
	
	\textbf{Noise processing and hidden layer objective.}
	To further bridge the gap between the label- and reference-based synthesis, we intend to link the implicit $R\sim N(0,I)$ in LEM and the explicit reference $X_r$ in REM. Particularly, we allocate a random $R$ for each training sample $X_r$ and make them into pairs $\left\{R,X_r \right\}$. During training, the pairs keep fixed. We use the constraint defined in (\ref{11}) for the optimization. 
	\begin{equation}
		\label{11}
		\begin{aligned}
			L_{sty} = \parallel S_{rand} - S_{ref} \parallel_1
		\end{aligned}
	\end{equation}
	Inspired by \cite{2019LOGAN,bond2020gradient}, we adopt a two-step strategy. In the first step, model parameters in LEM and REM are fixed, only $R$ gets updated. Then, the revised $R$ is given to LEM again for the new $S_{rand}$. The new $L_{sty}$ is computed to update the parameters in LEM and REM. This penalty allows LEM to learn different attribute styles and improve diversity. It also makes the conversion of REM more accurately.
	
	\textbf{Adversarial objective.}
	We employ the adversarial loss \cite{NIPS2014_5ca3e9b1} for the generation fidelity, formulated as (\ref{4}).
	\begin{equation}
		\label{4}
		\begin{aligned}
			L_{adv} = & \mathbb{E}_{X_s\sim p_{d}}[ \text{D}(X_s)]-\mathbb{E}_{X_g\in\{X_g^{l},X_g^{r},X_g^{i}\}}[\text{D}(X_g)]
		\end{aligned}
	\end{equation}
	Here $\text D$ is the discriminator constrained by 1-Lipschitz continuity following WGAN \cite{2017Wasserstein} and WGAN-GP \cite{NIPS2017_892c3b1c}. Besides $X_g^l$ and $X_g^r$, we randomly interpolate between $S_{rand}$ and $S_{ref}$, and give the results to $\text G$ for $X_g^{i}$,
	\begin{equation}
		\label{e5}
		\begin{aligned}
			X_g^{i}=\text{G}(X_s,(\alpha S_{rand} + (1-\alpha) S_{ref})), 
		\end{aligned}
	\end{equation}
	where $\alpha$ is a scalar and $\alpha\sim U(0,1)$.
	
	\textbf{Attribute classification.} We employ a classifier $\mathrm{C}$ 
	to ensure that generated images have accurate attributes \cite{2014Conditional}, formulated as (\ref{7}).
	\begin{equation}
		\label{7}
		\begin{aligned}
			L_{cls} = -\frac{1}{N_C} \sum_{i=0}^{N_C}[&y_i\log \mathrm{C}_i(X)+(1-y_i)\log(1-\mathrm{C}_i(X))]
		\end{aligned}
	\end{equation}
	Here $N_C$ is the number of attributes. $X$ is an image, including the translations $X_g^l$, $X_g^r$ and $X_g^i$, and the real image $X_s$.  $\mathrm{C}_i$ is the classifier that predicts the $i^{th}$ attribute of $X$. $y_i$ is the $i^{th}$ value of attribute label $Y$.
	
	\textbf{Source reconstruction.} We adopt the reconstruction loss in (\ref{8}) as a regularization.
	\begin{equation}
		\label{8}
		\begin{aligned}
			L_{rec} = &\parallel X_s - \mathrm{G}(X_s,\text{LEM}|\text{REM}(X_s,R,0)) \parallel_1
		\end{aligned}
	\end{equation}
	By setting $\mathrm{att}_{diff}=0$, it can ensure that the style information of the generated image comes from LEM or REM. 
	
	\textbf{Latent cycle consistency.}
	We employ a cycle consistency on the latent code $S_{rand}$ and $S_{ref}$, which can make $\mathrm{G}$ utilize the style code $S$ when generating $X_g$. The idea is to feed back the synthesis $X_g^l$ or $X_g^r$ as the reference input in REM, so that we can have the style code $\tilde{S}$ in (\ref{9}).
	\begin{equation}
		\label{9}
		\begin{aligned}
			L_{cyc} = &\parallel S_{rand} - \tilde{S}_{rand}
			\parallel_1
			+\parallel S_{ref} - \tilde{S}_{ref}
			\parallel_1,
		\end{aligned}
	\end{equation}
	Here $S_{rand}$ and $S_{ref}$ are the style codes for $X_g^l$ and $X_g^r$, respectively. Both $\tilde{S}_{rand}$ and $\tilde{S}_{ref}$ are computed from REM, to which we feed $X_s$ as the source input, and $X_g^l$ or $X_g^r$ as the reference. The $L_{cyc}$ in (\ref{9}) reflects the distance between the first and second time style code, which is similar to \cite{choi2020stargan,zhu2017multimodal}. Note that this penalty only affects $\mathrm{E}$ and $\mathrm{M}$, but not $\mathrm{G}$. 
	
	\textbf{Mode seeking objective.}
	To encourage the images with diverse styles, we use the mode seeking loss \cite{Mao_2019_CVPR,DBLP:journals/corr/abs-1901-09024} in (\ref{10}), 
	\begin{equation}
		\label{10}
		\begin{aligned}
			L_{ms} = \frac{1}{\parallel \text{G}(X_s, S_{rand}) -\text{G}(X_s,S'_{rand})\parallel_1}&\parallel R-R'\parallel_1,
		\end{aligned}
	\end{equation}
	where $R$ specifies $S_{rand}$ as (\ref{eq:eq2}), 
	and $R'\sim N(0,I)$ is a different input noise vector, giving $S'_{rand}$. 
	
	\textbf{Attribute keeping constraint.}
	In multi-attribute I2IT model, only the specified attributes need to be translated. The remaining ones are expected to be the same as the source. We use $\text{E}$ to extract features for unspecified attributes from both the original $X_s$ and edited $X_g^l$, constraining them to be close. 
	The formula is as follows.
	\begin{equation}
		\label{12}
		\begin{aligned}
			L_{ak} = 
			\parallel \text{E}(X_s,\mathrm{att}_{ak}^{Y_s\downarrow})
			- \text{E}(\text{G}(X_s,R,\mathrm{att}_{diff}^{Y_s\rightarrow Y_t}),\mathrm{att}_{ak}^{Y_s\downarrow}) \parallel_1
		\end{aligned}
	\end{equation}
	In (\ref{12}), we only extract the features that need to be retained, so we calculate $\mathrm{att}_{ak}^{Y_s\downarrow}\in\{-1,0,+1\}^n$ in (\ref{13}), in which attributes without editing from $Y_s$ to $Y_t$ are obtained.
	\begin{equation}
		\label{13}
		\begin{aligned}
			\mathrm{att}_{ak}^{Y_s\downarrow} = (1-2Y_s)(1-|\mathrm{att}_{diff}^{Y_s\rightarrow Y_t}|).
		\end{aligned}
	\end{equation} 
	\indent For example, if there are four attributes, given $Y_s=[1,0,1,0]$ and $Y_t=[1,1,0,0]$, we can obtain $\mathrm{att}_{diff}^{Y_s\rightarrow Y_t}=[0,1,-1,0]$ and 
	$\mathrm{att}_{ak}^{Y_s\downarrow}=[-1,0,0,1]$. In both $X_s$ and $X_g^l$, the features represented by the first and last attribute are extracted, and they are constrained to be close. The value -1 and 1 in $\mathrm{att}_{ak}^{Y_s\downarrow}$, corresponding to the attributes in $Y_s$ and $Y_t$ are 1 and 0, respectively. Please see the appendix, for a more detailed explanation of (\ref{13}).
	
	\textbf{Full objective.} 
	Finally, we train our $\text{M}$, $\text{E}$, $\text{G}$, $\text{D}$, $\text{C}$, and $r$ to minimize following objectives.
	\begin{align*}
		&L_{\mathrm{G}} = L_{adv} + \lambda_{cls}L_{cls} + \lambda_{rec}L_{rec} + \lambda_{sty}L_{sty} + \lambda_{ms}L_{ms}\\ 
		&+ \lambda_{ak}L_{ak}\\
		&L_{\mathrm{ME}} = L_{\mathrm{G}} + \lambda_{cyc}L_{cyc}\\
		&L_{\mathrm{DC}} = -L_{adv}+ \lambda_{cls}L_{cls} \qquad L_{r} = \lambda_{sty}L_{sty}
	\end{align*}
	where $\lambda_{cls}$, $\lambda_{rec}$, $\lambda_{cyc}$, $\lambda_{ms}$, $\lambda_{sty}$, and $\lambda_{ak}$ are hyperparameters for each term.
	\section{Experiments}
	\textbf{Datasets.} When converting multiple attributes simultaneously, a large amount of training data is needed. So, we adopt CelebA \cite{2014Deep} to evaluate our method. 
	Fourteen attributes are selected in our experiments, including Young, Mouth Slightly Open, Smiling, Black Hair, Blond Hair, Brown Hair, Gray Hair, Receding Hairline, Bangs, Male, No Beard, Mustache, Goatee, and Sideburns. Besides, 182,000 images are used as the training set, 
	and 19,962 as the testing set. We crop each image to 170$\times$170 and resize it to 128$\times$128. In order to prove the effectiveness of our method on the 
	single attribute and non-face data, we also evaluate it on the AFHQ published by StarGAN-V2, and follow StarGAN-V2's setting. In this part, there are three domains that need to be converted to each other, including cats, dogs, and wild animals. Each domain is provided with 5000 images. These images are resized to 256$\times$256 resolution for training.
	
	\indent\textbf{Evaluation metrics.} We use Frechét inception distance (FID) \cite{NIPS2017_8a1d6947} and Inception Score (IS)\cite{salimans2016improved} to evaluate the visual quality, and evaluate the diversity of generated images by the learned perceptual image patch similarity (LPIPS) \cite{Zhang_2018_CVPR,2012ImageNet}. Besides, 
	the domain Accuracy of the generated image is evaluated by a pre-trianed multi-attribute classifier. We compute metrics for all attributes on the test data and report their averages. Please see the appendix for the accuracy of each attribute and other implementation details. 
	\subsection{Qualitative and Quantitative Results}
	\begin{figure*}[ht]
		\centering
		\includegraphics[scale=0.62]{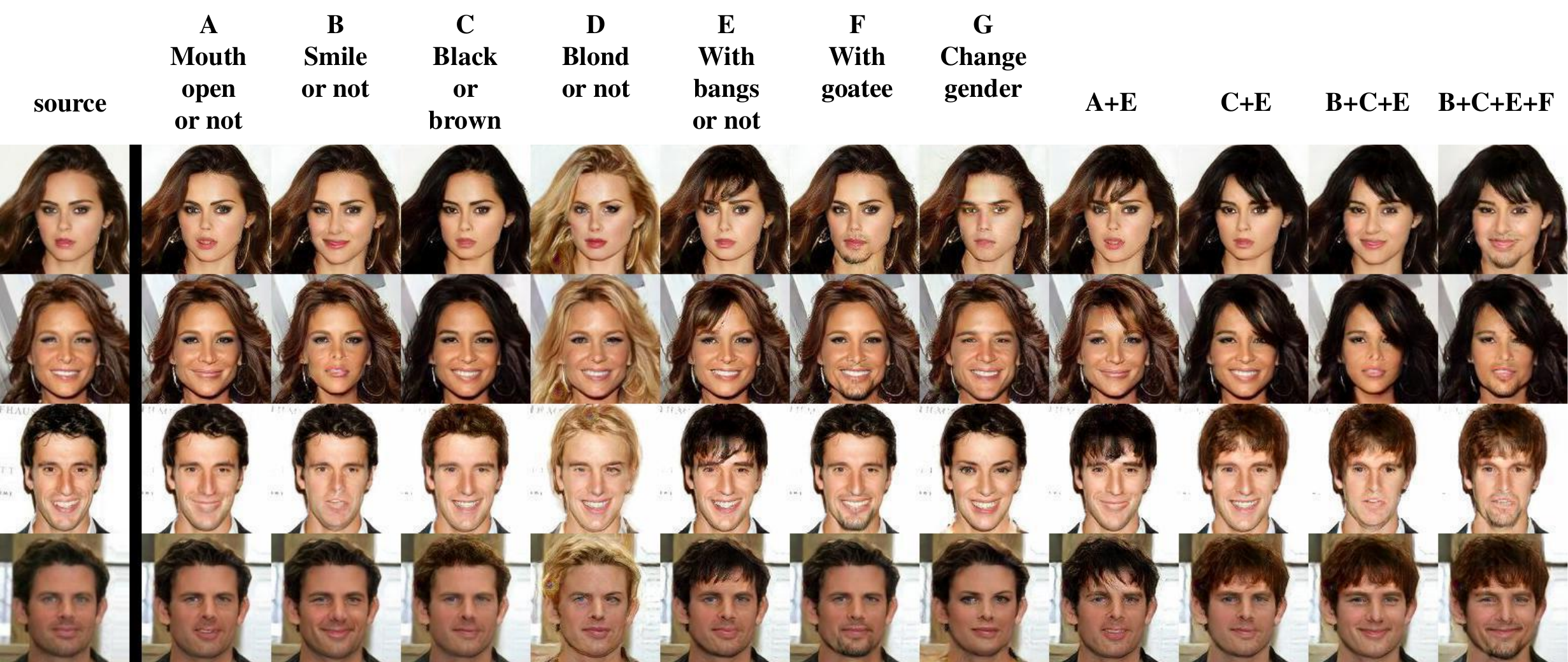}
		\caption{\textbf{Label-based synthesis of our model.} The last 4 columns are results by editing more than one attributes.}
		\label{fig:fig3}
	\end{figure*}
	\begin{figure*}[ht]
		\centering
		\includegraphics[width=0.9\textwidth]{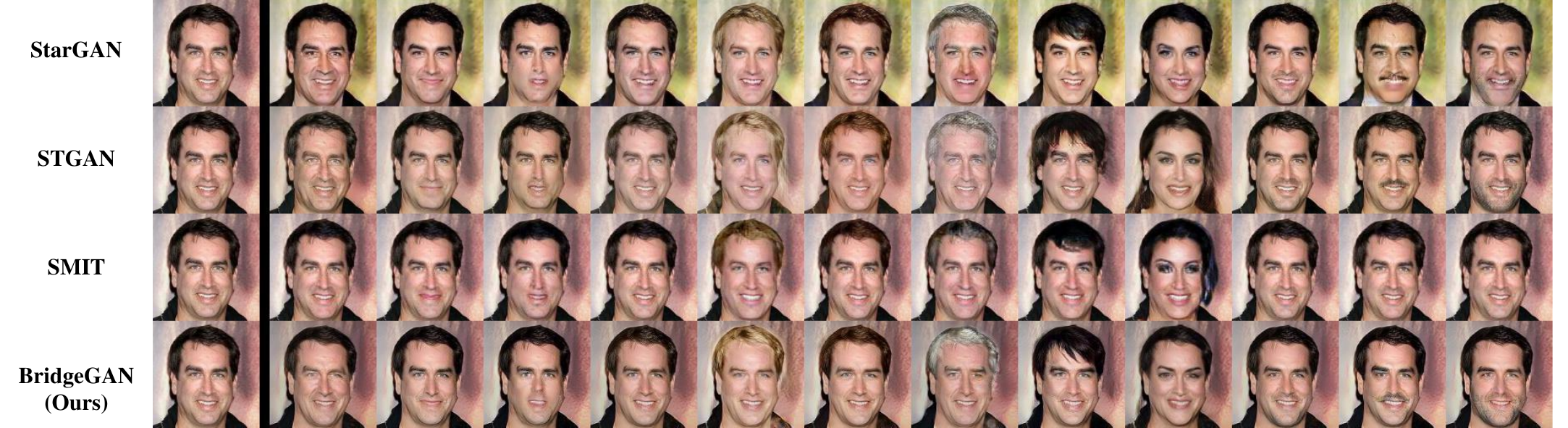}
		\caption{\textbf{Label-based synthesis of 4 models.} From left to right: source, young, mouth slightly open, smiling, black, blond, brown, gray hair, bangs, male, no beard, mustache, and sideburns.}
		\label{fig:fig4}
	\end{figure*}
	\begin{figure*}[ht]
		\centering
		\includegraphics[scale=0.6]{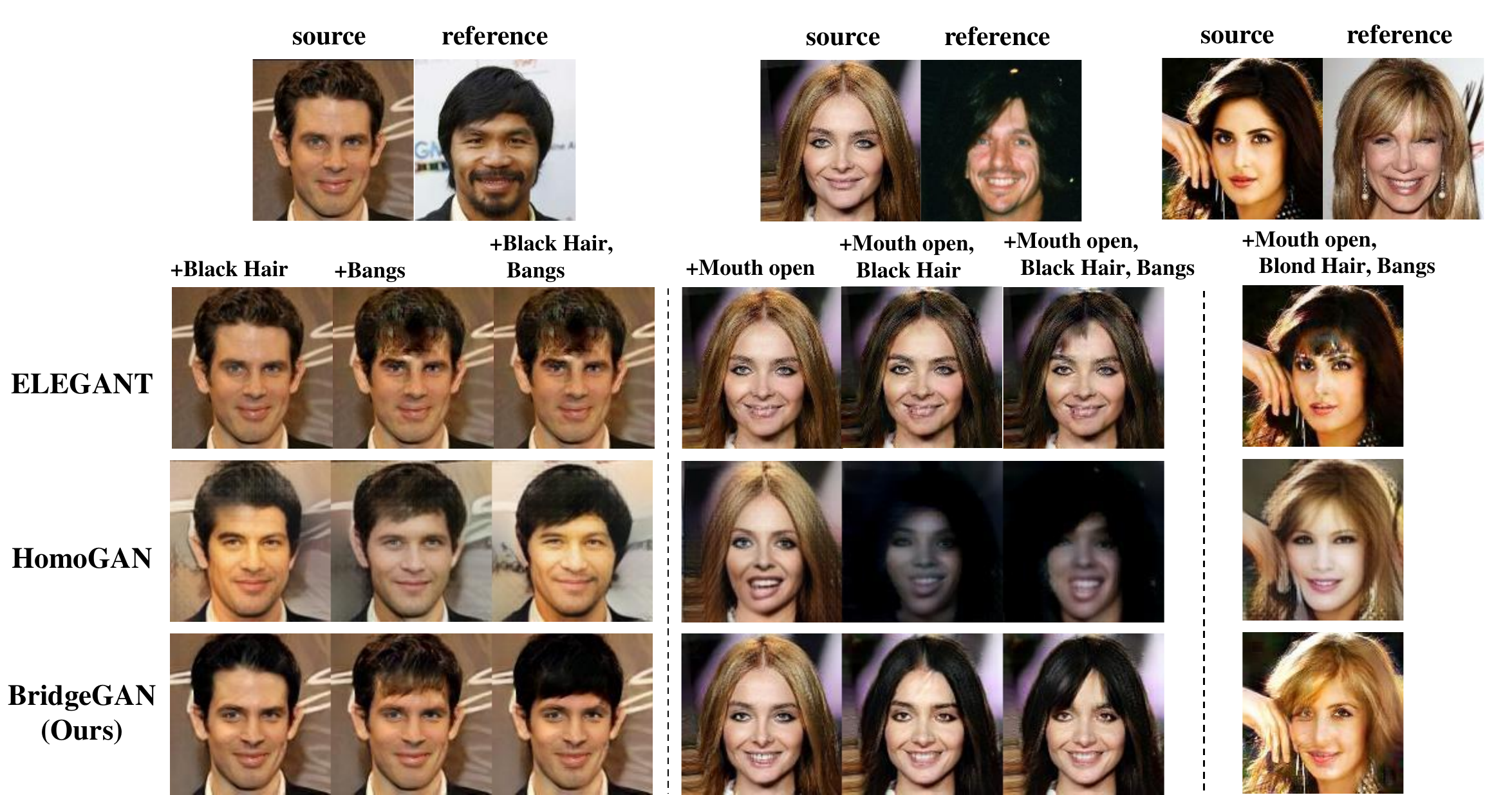}
		\caption{\textbf{Reference-based synthesis results.} We show 3 data pairs, and their results on single or multiple attributes editing.}
		\label{fig:fig5}
	\end{figure*}
	
	\begin{figure*}[ht]
		\centering
		\includegraphics[scale=0.6]{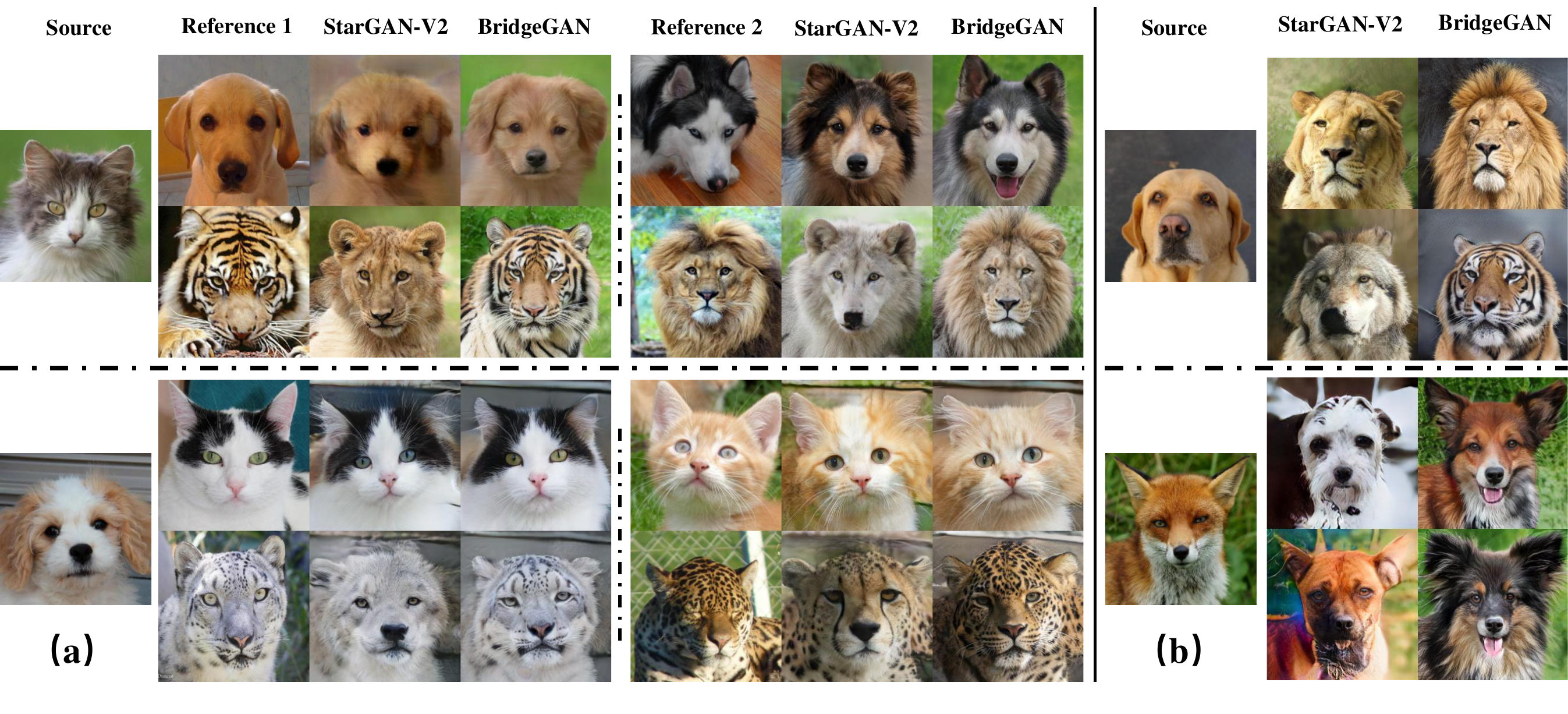}
		\caption{\textbf{Qualitative comparison of reference-based and label-based synthesis results on the AFHQ datasets.} On the left, we show the reference-based synthesis of different methods, each source image is equipped with 4 reference images with different styles. On the right, we show the label-based synthesis results, including dog-to-wild, and wild-to-dog. Each source image is equipped with two different sampling noise vectors. Please zoom in for more details.}
		\label{fig:afhq}
	\end{figure*}
	
	\indent\textbf{Label-based synthesis. }
	Fig.\ref{fig:fig3} shows representative examples, demonstrating that our method can generate high-quality images, and accurately control the attributes to translate. Moreover, the 9$th$ to 12$th$ columns in Fig.\ref{fig:fig3} are results of simultaneous conversion of multiple attributes. We find it does not degrade the image quality and change the irrelevant attributes. Fig.\ref{fig:fig4} shows a visual comparison among 4 models, including StarGAN, STGAN, SMIT and ours. Note that since label-based synthesis is relatively easy, all models accomplish the task. However, StarGAN seriously changes the background. Compared with our results, STGAN cannot accurately edit certain attributes while keeping others unchanged. \emph{E.g.}, when changing the hair color, the eyebrow color is altered obviously. When editing the gender, the hair style changes notably. SMIT cannot edit on beards, and the quality for gender conversion is poor.
	
	Tab.\ref{tab:1} lists FID, Accuracy, and LPIPS of all competing methods. The performance of StarGAN, STGAN, and SMIT is obviously worse than our method (model G). As we all know, for label-based synthesis, the diversity and attribute accuracy of the generated images are conflicting to a certain extent. Few works can pursue both at the same time except ours. In fact, StarGAN-V2 achieve good results in face conversion, but it cannot complete the task of multi-attribute conversion, so we cannot compare with it.
	\begin{figure}[ht]
		\centering
		\includegraphics[scale=0.40]{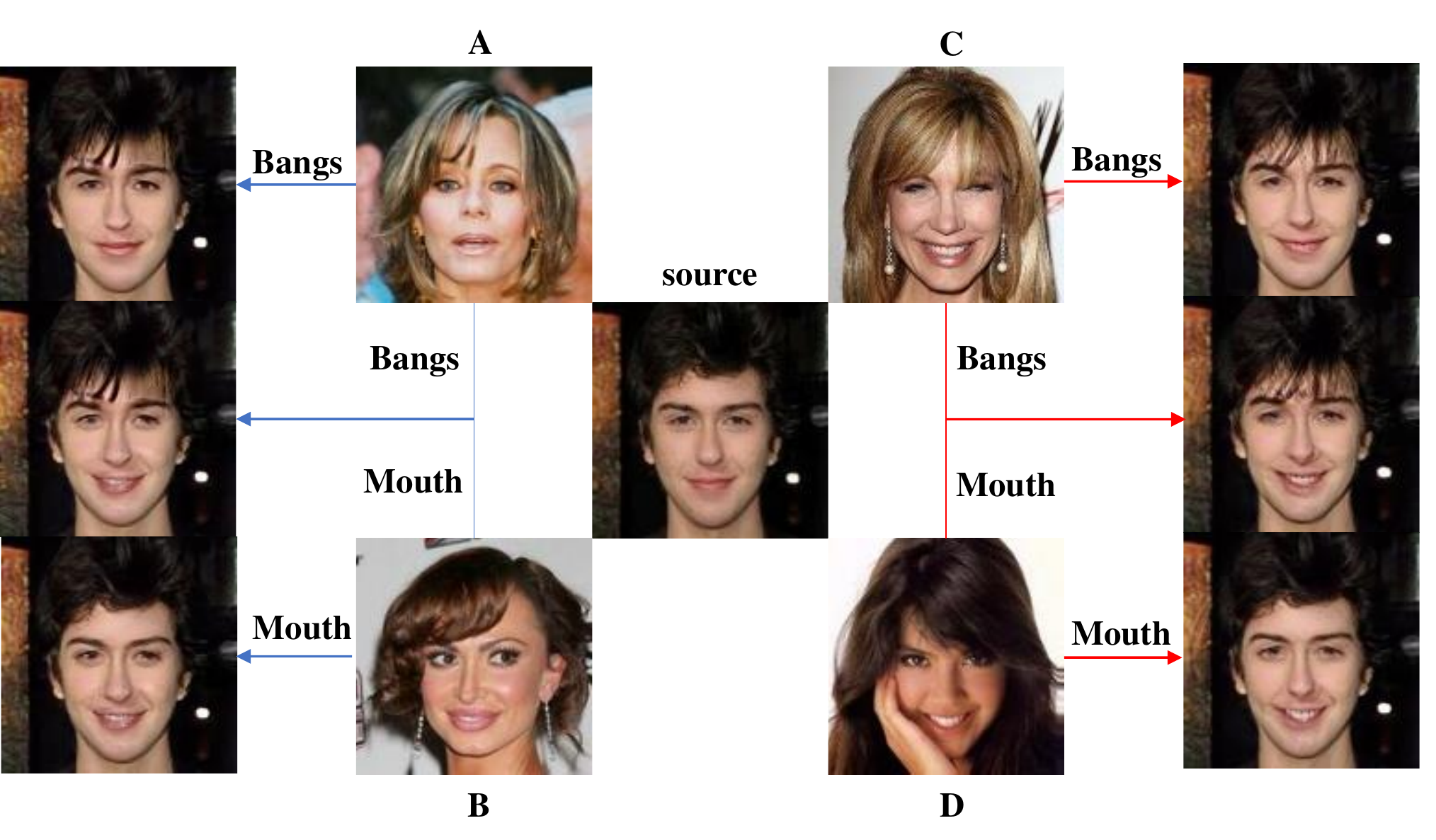}
		\caption{\textbf{Reference-based multi-attribute editing by latent code averaging on $S_r^r$.} The results on the left are provided by reference image A and B, while the right are provided by C and D. }
		\label{fig:fig6}
	\end{figure}
	\begin{figure}[ht]
		\centering
		\includegraphics[scale=0.47]{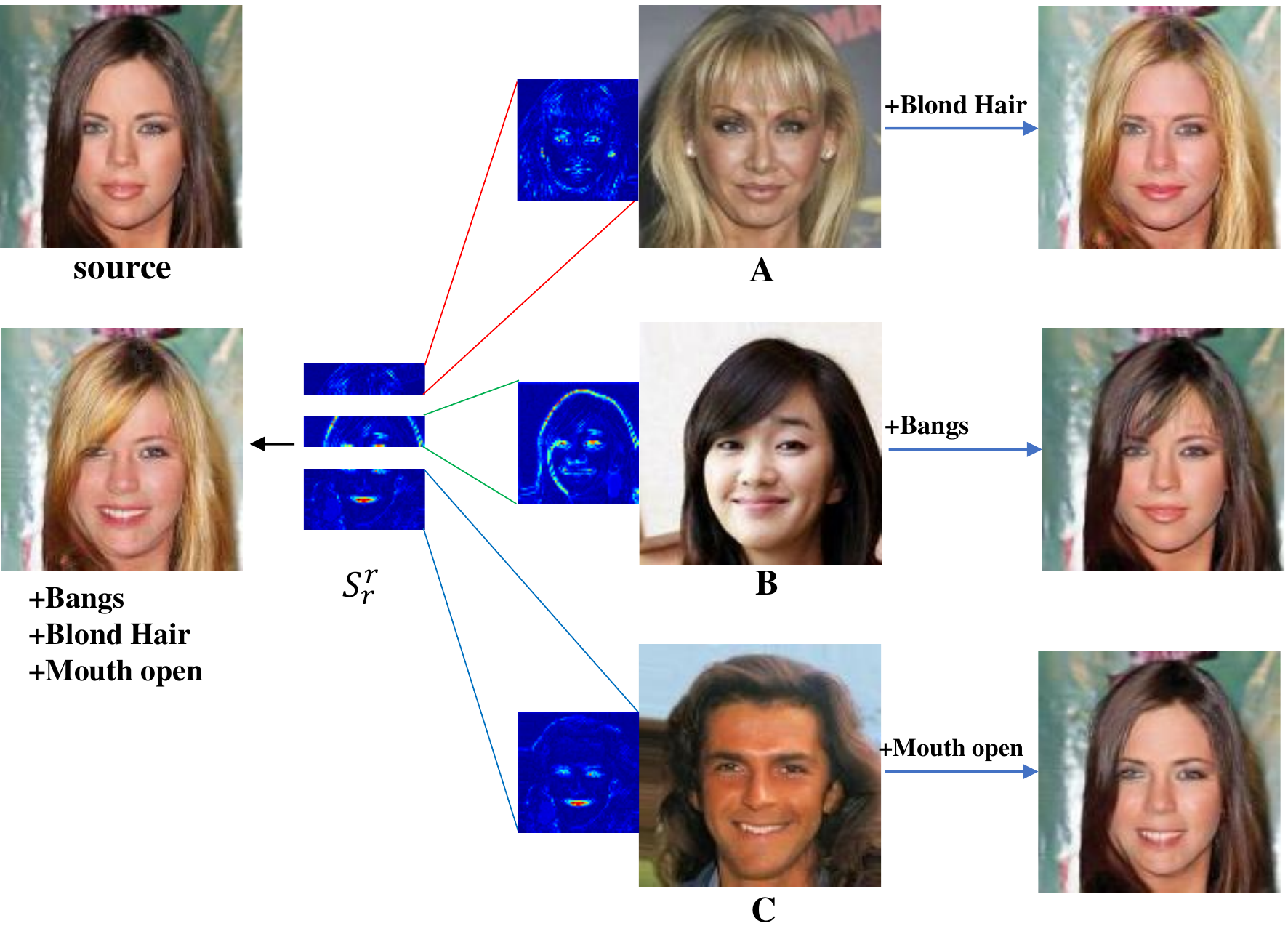}
		\caption{\textbf{Reference-based multi-attribute editing by latent code mixing.} On the left, we mix $S_r^r$ from 3 different references (A: blond hair, B: bangs, C: mouth open). On the right, we take a single reference and make the translation.}
		\label{fig:fig7}
	\end{figure}
	
	\begin{table}[ht]
		\begin{center}
			\scalebox{0.9}{
				\setlength{\tabcolsep}{1.5mm}{
					\begin{tabular}{ l l l c| c| c}
						\hline
						&method && FID$\downarrow$ &Accuracy$\uparrow$ &LPIPS$\uparrow$\\
						\hline
						&StarGAN \cite{choi2018stargan}&&25.96$|$-\qquad\qquad&69.5$|$-\qquad\!\!&-\\
						&STGAN \cite{liu2019stgan}&&16.11$|$-\qquad\qquad&80.6$|$-\qquad\!\!&-\\
						&SMIT \cite{romero2019smit}&&\textbf{12.14}$|$-\qquad\qquad&27.4$|$-\qquad\!\!&0.029\\
						&ELEGANT \cite{xiao2018elegant}&& \qquad-$|$68.88& \ \quad-$|$26.6&-\\
						&HomoGAN \cite{chen2019homomorphic}&& \qquad-$|$17.96& \ \quad-$|$30.5&-\\ 
						\hline
						&A:OLEM&&25.10$|$-\qquad\qquad&84.3$|$-\qquad\!\!\! &0.020 \\
						&B:+OREM&&17.07$|$13.91&83.5$|$47.6&0.017 \\
						&C:+$S_s$&&\!\!\!\textbf{12.14}$|$9.52 &82.8$|$47.5&0.006 \\
						&D:+interp&&17.32$|$10.09&86.2$|$42.1&0.012 \\
						&E:+$L_{ms}$&&\!\!\!14.03$|$\textbf{9.28}&82.9$|$42.7&0.030 \\
						&F:+$L_{sty}$&&16.82$|$12.95&85.1$|$39.5&0.034 \\
						&G:+$L_{ak}$&&14.26$|$11.38&\textbf{87.0}$|$\textbf{54.0}&\textbf{0.043} \\    
						\hline
					\end{tabular}
				}
			}
		\end{center}
		\caption{\textbf{Quantitative comparisons and ablation studies by the metrics.} For FID and Accuracy we measure them on two types of synthesis. On the left of the separator $|$ is the value of label-based synthesis, while the right side is reference-based. For LPIPS, we only evaluate on the probabilistic models by random sampling on the noise input.}
		\label{tab:1}
	\end{table}

	\begin{table}[ht]
		\begin{center}
			\scalebox{0.9}{
				\setlength{\tabcolsep}{1.5mm}{
					\begin{tabular}{ l l l c| c| c}
						\hline
						&method && FID$\downarrow$&LPIPS$\uparrow$&IS$\uparrow$\\
						\hline
						&StarGAN-V2 \cite{choi2020stargan}&& 31.07$|$33.32&0.478$|$\textbf{0.437}\!\!&4.112\\
						&BridgeGAN&&\textbf{25.87}$|$\textbf{26.68}&\ \textbf{0.479}$|$0.432&\textbf{4.666}\\
						\hline
				\end{tabular}}
			}
		\end{center}
		\caption{\textbf{Quantitative comparisons on the AFHQ datasets by the metrics.} We measure these on two types of synthesis, the result of IS is the average of them. 
			In order to measure the image quality in general, we use the real images of the test set to calculate the FID.}
		\label{tab:2}
	\end{table}
	
	\begin{figure}[ht]
		\centering
		\includegraphics[scale=0.5]{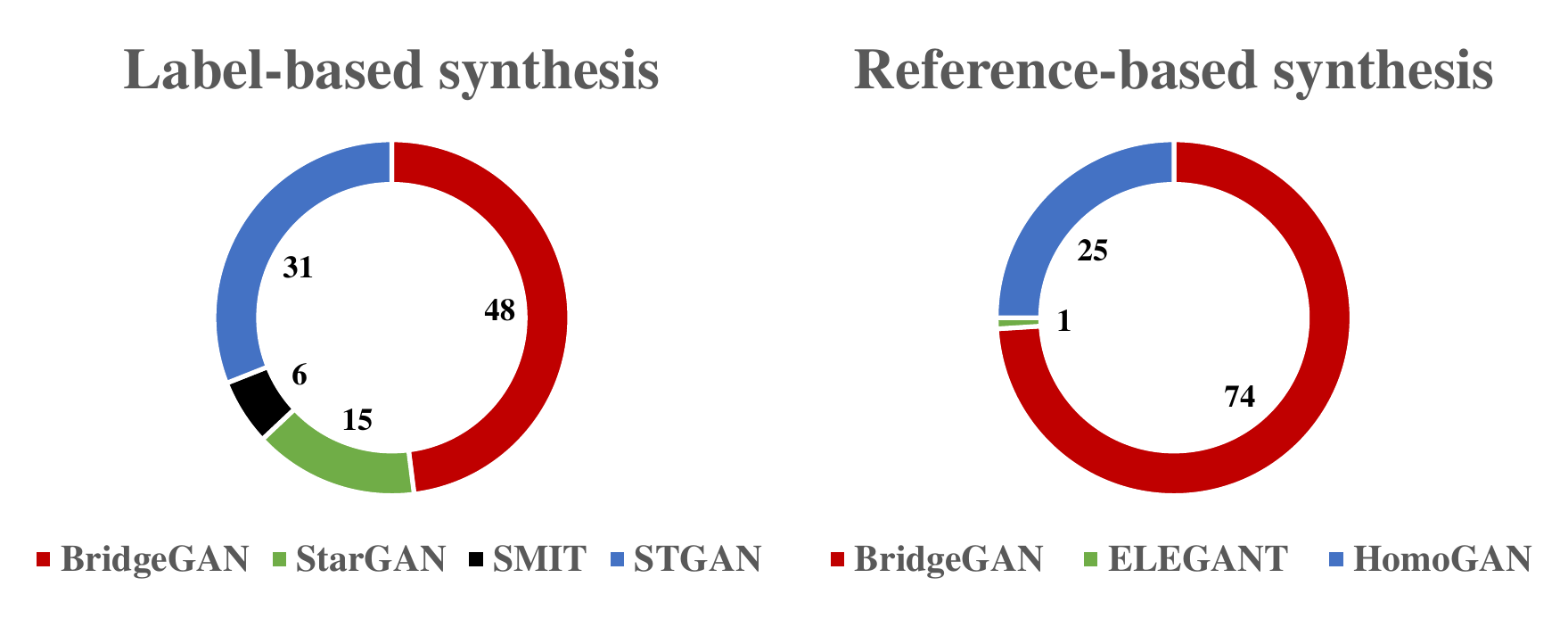}
		\caption{\textbf{User study results of two types of synthesis.}}
		\label{fig:fig10}
	\end{figure}
	\textbf{Reference-based synthesis.}
	In Fig.\ref{fig:fig5}, we visually compare the results of HomoGAN, ELEGANT and ours. Obviously, HomoGAN achieves the low quality, and cannot keep irrelevant factors such as background and skin color. ELEGANT has high quality but often fails to change appropriately. In our model, $\text{LEM}$ can assist $\text{REM}$ to locate the relevant attributes more accurately. Tab.\ref{tab:1} also lists the quantitative metrics, including FID and Accuracy. Our model outperforms the HomoGAN and ELEGANT undoubtedly on these two metrics.
	
	We further test our model on the synthesis from multiple references. Particularly, we use the encoder $\text{E}$ to extract the features $S_r^r$ from different references, and average on $S_r^r$ to give the result, as is shown in Fig.\ref{fig:fig6}. The model succeeds taking the relevant attributes from different references. In addition, we also perform the feature mixing on $S_r^r$. Here, $S_r^r$ is mixed along the image height by the original codes from different references, and the results is shown in Fig.\ref{fig:fig7}. The model also takes the relevant attributes and their styles from corresponding references.
	
	\indent\textbf{User Study.} In Fig.\ref{fig:fig10}, we conduct a user study to evaluate different models under human perception. For label-based synthesis, we randomly choose images to translate for each attribute. Users are asked to select the best editing result from all competing methods. For reference-based synthesis, we randomly generate translated results . Users are required to choose the best among all competing methods, according to whether the converted attributes between the synthesis and the reference image are similar, and whether other irrelevant attributes remain the same.
	
	\indent\textbf{Experiments on single attribute.} In Fig.\ref{fig:afhq}, we show the visually results on the AFHQ datasets. So far, StarGAN-V2 seems to be the best in the AHFQ, so we only compared with it. In comparison, we are better than StarGAN-V2 in maintaining the background information of the original image. In terms of the similarity of the reference image, our results show a higher degree of stability. In addition, Tab.\ref{tab:2} lists FID, IS and LPIPS of all methods. On the one hand, we achieved lower FID score and higher IS in the quality of images. On the other hand, with similar LPIPS scores, we can maintain background information to a certain extent. This proves that our method can capture the characteristics of domains and edit more accurately.
	
	\indent\textbf{Interpolation results.}In Fig.\ref{fig:fig1}, we use the method in Eq.(\ref{e5}) to synthesize the interpolation between the two types of latent codes, $S_{rand}$ and $S_{ref}$. The interpolated images between the two methods are natural and achieve smooth transitions. Our model has the ability to give diverse results with the specific domain style linearly approaching to the reference.
	
	\subsection{Ablation Study} \label{ablation}
	Tab.\ref{tab:1} lists the metrics for ablation study. The baseline is the model A, in which the network has only the LEM (OLEM). Therefore, we remove the whole $\text{REM}$ and the $S_s$ in $\text{LEM}$, which means $S_{rand} = S_r^l$. We use $L_{adv}$ in (\ref{4}), $L_{cls}$ in (\ref{7}) and $L_{rec}$ (\ref{8}) for training. Intuitively, we can also have a model with only the REM (OREM) for reference-base synthesis. But such a model can not take the source content.
	
	Then we add individual component to A. In model B, we combine the OLEM with the OREM, and add $L_{cyc}$ into the objectives. Obviously, once we start to establish the connection between $\text{LEM}$ and $\text{REM}$, the reference-based result becomes better. Then we add the encoder $\mathrm{E}$ to give $S_s$ in model C. This means that the domain difference between the original source and the reference is constructed by $S_{rand}$ and $S_{ref}$ in the hidden layer. The metric of FID shows it greatly improves image quality while keeping others the same. We further add the interpolation image $X_g^i$ to $L_{adv}$ for training in model D. It increases the Accuracy and LPIPS for label-based synthesis. For diverse results, we add $L_{ms}$ in (\ref{10}) in the setting E, and LPIPS for label-based synthesis is improved greatly. Moreover, this loss has a good impact on the reference-based synthesis. We then add $L_{sty}$ in (\ref{11}) to model F to give a tight link between LEM and REM. Compared to method E, Accuracy and LPIPS of Label-based synthesis are further improved. Finally, we add $L_{ak}$ (\ref{12}) in the last model G, which aims to keep irrelevant attributes from being converted. Note that it can guide both $\text{LEM}$ and $\text{REM}$, particularly it gives a higher Accuracy. 
	
	\section{Conclusion}
	This paper constructs a new architecture for multi-attribute I2I translation. Our model bridges the gap between label- and reference-based synthesis, so that both of them get improved. For the label-based image synthesis, we can simultaneously obtain diverse and high-accuracy translated images. For the reference-based synthesis, our model is able to take the specific style similar to the reference. The results show that the proposed model remarkably outperforms the previous ones.
	{\small
		\bibliographystyle{ieee_fullname}

	}
	
\end{document}